%% file: 2022_emnlp_cheater.tex
\pdfoutput=1

\documentclass[11pt]{article}

\usepackage[]{style/emnlp2022}

\usepackage{times}
\usepackage{latexsym}
\usepackage{subfigure}
\usepackage[T1]{fontenc}

\usepackage[utf8]{inputenc}

\usepackage{microtype}
\usepackage{mathtools}
\usepackage{inconsolata}
\usepackage{graphicx}
\usepackage{xurl}
\usepackage[ruled,vlined]{algorithm2e} 
\usepackage{algorithmicx}
\usepackage{multirow}
\usepackage{booktabs}
\usepackage{thmtools} 
\usepackage{thm-restate}
\usepackage{amsthm}
\usepackage{bbm}
\usepackage{quoting}
\quotingsetup{vskip=0pt}

\newif\ifcomment\commentfalse
\input{style/preamble}

\usepackage{color}
\definecolor{deepblue}{RGB}{0, 51, 204}

\title{
Cheater's Bowl: Human vs. Computer Search Strategies for Open-Domain Question Answering
}

\author{
Wanrong He \\
Tsinghua University\thanks{\ \ Work completed at University of Maryland.}\\
\emaillink{hewanrong8@gmail.com}
  \And
Andrew Mao \\
University of Maryland \\
\emaillink{amao1@terpmail.umd.edu}
  \And
Jordan Boyd-Graber\\
University of Maryland \\
\emaillink{jbg@umiacs.umd.edu}
}
\date{}

\begin{document}
\maketitle

\begin{abstract}
    \input{sections/00-abstract}
\end{abstract}

\input{sections/10-intro}
\input{sections/20-search}
\input{sections/30-data-collection}

\input{sections/40-analysis}
\input{sections/50-model}
\input{sections/60-related}

\input{sections/70-conclusion}

\input{sections/acks}


\bibliography{bib/journal-full,bib/anthology,bib/wanrong,bib/jbg}
\bibliographystyle{style/acl_natbib}

\clearpage
\begin{appendix}
    \input{sections/apx}
\end{appendix}

\end{document}

%% file: style/preamble.tex
\usepackage[a-1b]{pdfx}

\usepackage{framed}
\usepackage{mdwlist}
\usepackage{siunitx}
\usepackage{latexsym}
\usepackage{colortbl}
\usepackage{xcolor}
\usepackage{nicefrac}
\usepackage{booktabs}
\usepackage{fnpct}
\usepackage{amsfonts}
\usepackage[T1]{fontenc}
\usepackage{bold-extra}
\usepackage{amsmath}
\usepackage{amssymb}
\usepackage{bm}
\usepackage{graphicx}
\usepackage{mathtools}
\usepackage{microtype}
\usepackage{multirow}
\usepackage{multicol}
\usepackage{xpatch}
\usepackage{latexsym,comment}
\usepackage[normalem]{ulem}

\newcommand*{\missingreference}{{\Huge \colorbox{red}{?reference?}}}
\newcommand*{\missingcitation}{{\Huge \colorbox{red}{?citation?}}}

\makeatletter
\xpatchcmd{\@setref}{\bfseries}{\missingreference}{}{}
\def\@citex[#1]#2{\leavevmode
    \let\@citea\@empty
    \@cite{\@for\@citeb:=#2\do
        {\@citea\def\@citea{,\penalty\@m\ }%
            \edef\@citeb{\expandafter\@firstofone\@citeb\@empty}%
            \if@filesw\immediate\write\@auxout{\string\citation{\@citeb}}\fi
            \@ifundefined{b@\@citeb}{\hbox{\reset@font\missingcitation}%
                \G@refundefinedtrue
                \@latex@warning
                {Citation `\@citeb' on page \thepage \space undefined}}%
            {\@cite@ofmt{\csname b@\@citeb\endcsname}}}}{#1}}
\makeatother

\newcommand{\gem}[1]{\mbox{\textsc{gem}}}
\newcommand{\abr}[1]{\textsc{#1}}

\renewenvironment{quote}
{\list{}{\rightmargin\leftmargin}%
    \item\relax\small\ignorespaces}
{\unskip\unskip\endlist}

\newcommand{\emaillink}[1]{{\small \href{mailto://#1}{\texttt{#1}}}}

\newcommand{\bmt}{\abr{bm}{\small 25}}

\newcommand{\hidetext}[1]{}
\newcommand{\ignore}[1]{}

\ifcomment
    \newcommand{\pinaforecomment}[3]{\colorbox{#1}{\parbox{.8\linewidth}{#2: #3}}}

    \newcommand{\prtodo}[1]{\pinaforecomment{lightblue}{pr}{#1}}
    \newcommand{\prtodoi}[1]{\pinaforecomment{lightblue}{pr}{#1}}
\else
    \newcommand{\pinaforecomment}[3]{}
    \newcommand{\prtodo}[1]{}
    \newcommand{\prtodoi}[1]{}
\fi

\newcommand{\jbgcomment}[1]{\pinaforecomment{red}{JBG}{#1}}

\newcommand{\smallurl}[1]{ \begin{tiny}\url{#1}\end{tiny}}

\definecolor{lightblue}{HTML}{3cc7ea}
\definecolor{CUgold}{HTML}{CFB87C}
\definecolor{grey}{rgb}{0.95,0.95,0.95}
\definecolor{ceil}{rgb}{0.57, 0.63, 0.81}
\definecolor{UMDred}{HTML}{ed1c24}
\definecolor{UMDyellow}{HTML}{ffc20e}

\newcommand{\qb}[0]{\abr{qb}}
\newcommand{\qa}[0]{\abr{qa}}

\newcommand{\hotpot}{\abr{h}{\small otpot}\abr{qa}}

%% file: sections/00-abstract.tex

For humans and computers, the first step in answering an open-domain
question is retrieving a set of relevant documents from a large
corpus.
However, the strategies that computers use fundamentally
differ from those of humans.
To better understand these differences, we design a gamified interface
for data collection---Cheater's Bowl---where a human answers complex
questions with access to both traditional and modern search tools.
We collect a dataset of human search sessions, analyze human search
strategies, and compare them to state-of-the-art multi-hop \qa{} models.
Humans query logically, apply dynamic search chains, and
use world knowledge to boost searching.
We demonstrate how human queries can improve the accuracy of existing
systems and propose improving the future design of \qa{} models.

%% file: sections/10-intro.tex
\section{The Joy of Search: Only for Humans?}

\jbgcomment{Get rid of the cite dump, }

A grand goal of artificial intelligence research is to design agents
that can search for information to answer complex questions. Modern
day question answering (\qa{}) models have the ability to issue
text-based queries to a search engine~\cite{qi-etal-2019-answering,
  qi-etal-2021-answering, xiongAnsweringComplexOpenDomain2020,
  zhao-etal-2021-multi-step, adolphsBoostingSearchEngines2021,
  nakanoWebGPTBrowserassistedQuestionanswering2021} and use multiple
iterations of querying and reading to search for an answer.
However, there is still a performance gap between machines and humans.

%
%




%
Dan Russell describes humans with virtuosic search abilities in his book
\textit{The Joy of Search}~\cite{russellJoySearchGoogle2019}, and
describes search strategies that: use world knowledge;
use parallel search chains, abandon futile threads;
and use multiple sources and languages.
%
While we can all admire Dan Russell's search skills, it does
not answer the question: how different are computers' searches from those of humans?


This paper attempts to answer this question with a 
comparison of human and computer search strategies.
We create
``Cheater's Bowl'', an interface that gamifies answering questions, with
the addition of tools such as a traditional search engine, a neural
search engine, and modern \qa{} models.
We collect a dataset of human search sessions while using our
interface to answer complex open-domain multi-hop questions
(Section~\ref{collect}) from
Quizbowl~\cite[\qb{}]{rodriguezQuizbowlCaseIncremental2021} and
HotpotQA~\cite{yang-etal-2018-hotpotqa}.
We analyze the differences between human and computer search
strategies and detail where current models fall short
(Section~\ref{analysis}).
Substituting queries generated by models with human queries significantly improves model accuracy. 
We propose design suggestions
for future query-driven \qa{} models, such as creating retriever-aware queries and validating answers. Our dataset can serve as the foundation
for training them (Section~\ref{model}).

Our main contributions are:
\begin{itemize*}
\item We create an interface for answering questions with access to search tools.
    \item We collect a dataset of human search sessions on questions
      from Quizbowl and HotpotQA.
    \item We compare human and computer strategies for \qa{}: humans apply dynamic search chains, use world knowledge, and
      reason logically. We propose these as potential directions for
      query-driven \qa{} models.
\end{itemize*}

%% file: sections/20-search.tex
\section{How Humans and Computers Search}
\label{sec:search}

To compare how humans and computers form queries to answer questions,
we first need to have a level playing field and set up our
vocabulary.
Sometimes, we will need to speak abstractly about who is trying to
answer the question without distinguishing between the human and the
computer.
In these cases, we refer to them as an ``agent'', which can be either
the human or the computer.
We assume that the agents do not know the answers directly and that
they create text-based queries to find the answer (we discuss the
alternatives, closed book \abr{qa}, directly forming dense queries and other computer systems,
in Section~\ref{sec:alt}).


We assume that agents, given an initial question, form a
text query~$q_0$.
The $i$\textsuperscript{th} query $q_i$ retrieves a set of documents~$\mathcal{D}_{i+1}
= \{d_{1}, \ldots, d_{|\mathcal{D}_{i+1}|}\}$ from a large corpus of
documents~$\mathcal{D}$---in our case all the paragraphs in
Wikipedia pages.
The retrieved documents provide additional information, allowing the
agent to answer the question or compose a new query~$q_{i+1}$.
The set of documents~$\mathcal{E}_{i} \subseteq \mathcal{D}_{i}$
provide information---evidence---for answering the question with
answer~$a$ or composing subsequent queries~$\{q_j|j> i\}$.
It is possible that not all retrieved documents are
read---$\mathcal{E}_{i} \neq \mathcal{D}_{i}$---since not all of the
retrieved documents are relevant, and an agent
might only read a few of them.
This process repeats until the agent answers the question.
We represent the iterative question-answering process as action path:
$A= (q_0, \mathcal{E}_{1}, q_1, \mathcal{E}_{2}, q_2, \cdots,
\mathcal{E}_{k}, a)$.

\subsection{Human Queries}

How humans form queries when they search for an answer depends on many
factors, as summarized by \citet{allen-91:topic}: the experience of
the user searching for information, how much the user knows about the
topic, and whether they are finding completely new information or
navigating to a specific information source they have seen before.
Beyond the intrinsic knowledge of particular users, users often have
particular strategies they favor. 
For example, users may copy/paste information into a document, keep
multiple tabs open, or always turn to a particular source of
information first~\cite{aula-05:search}.

\subsection{Computer Systems}

Thanks to the recent development of machine learning and natural
language understanding, computer systems
can answer open-domain questions by generating text-based
queries.
The \abr{golden} retriever~\cite{qi-etal-2019-answering} generates a
query~$q_k$ at reasoning step~$k$ by selecting a substring from the
current reasoning path~$R_k$, which is the concatenation of the
question $Q$ and previously selected retrieval results at each
reasoning step: $R_k = (Q, d_1, d_2, \cdots, d_{k})$, $R_0=(Q)$ (for questions with $n\geq 1$ clues/sentences, we use their
concatenation as the full question $Q = (Q_0, Q_1, \cdots,
Q_{n-1})$).
\abr{golden} retriever then selects a single document $d_{k+1}$ from the set
of documents $\mathcal{D}_{k+1}$ retrieved by $q_k$, appends~$d_{k+1}$
to the current reasoning path and forms an updated reasoning
path~$R_{k+1}$.
\abr{irrr}~\cite{qi-etal-2021-answering} further advances
\abr{golden} retriever by allowing queries to be any subsequence of the
reasoning path, though less flexible than human queries.
At each step, these systems only select a document as evidence
for further actions: $\mathcal{E}_i = \{d_i\}$.
Thus the action path becomes
$A = (q_0, \mathcal{E}_{1}, q_1, \mathcal{E}_{2}, q_2, \cdots,
\mathcal{E}_{k}, a) = \left(q_{0}, \{d_1\}, q_{1}, \{d_2\}, q_{2},
  \cdots, \{d_k\}, a\right)$.

\jbgcomment{Explain the last sentence} 


%% file: sections/30-data-collection.tex
\section{Cheater's Bowl: Gamified Data Collection For Human Searches}\label{collect}

This section discusses the gamified data collection process via an
example.

\subsection{Motivation}

High-stakes trivia competitions test who knows
more about a particular topic.
However, it has occasionally been plagued by cheater
scandals from Charles van Doren in the 1950s~\cite{tedlow-76:scandals}
to Andy Watkins in the 2010s~\cite{trotter-13:harvard}.
These scandals are not particularly relevant to computational
linguistics, but the move to online trivia competitions during the
Corona pandemic brought a new form of cheating to the fore: people
would see a trivia question and quickly try to use a search engine to
find the answer.

\begin{figure*}[t]
    \centering
    \includegraphics[width=0.8\linewidth]{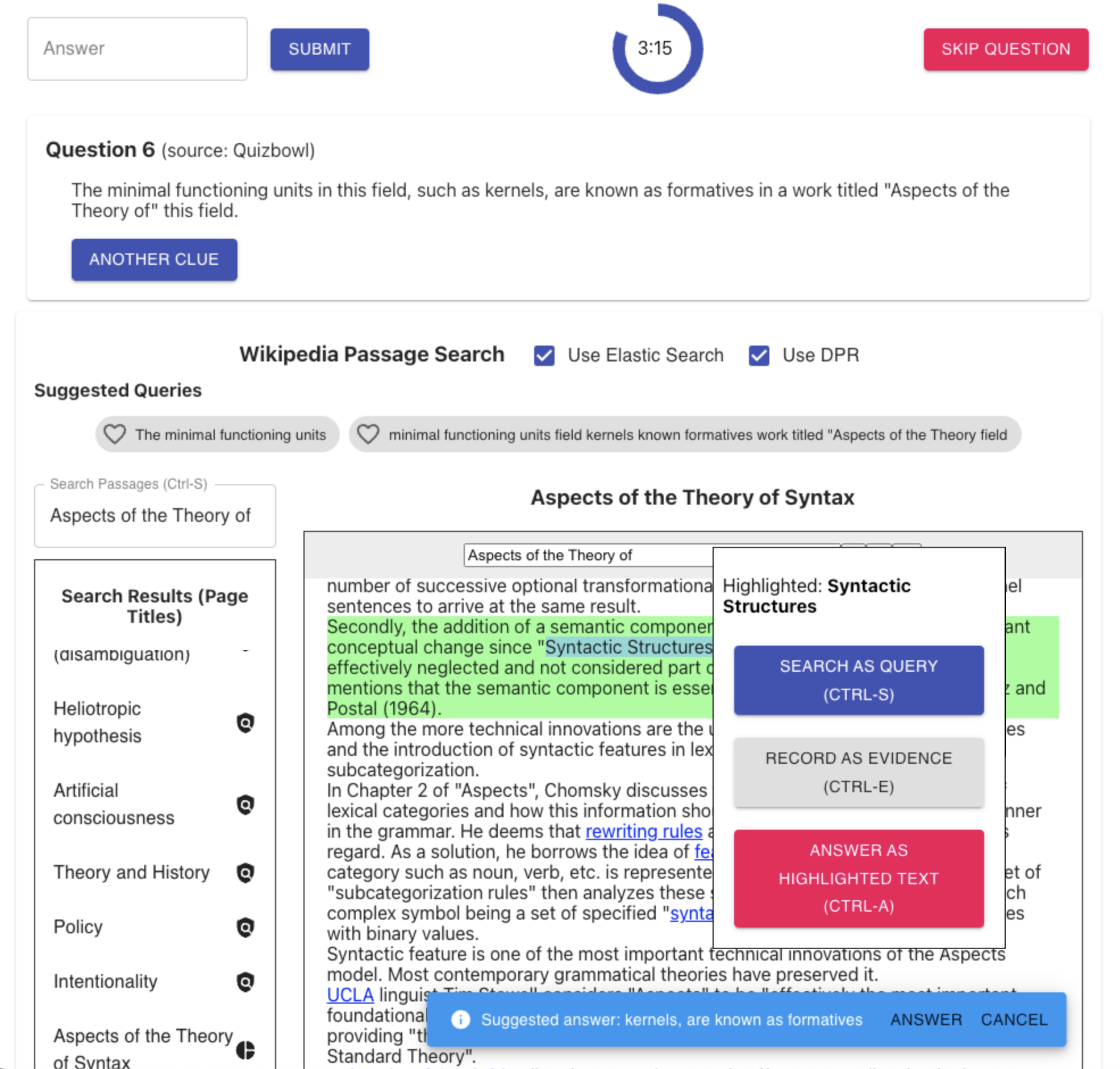}
    \caption{User interface for Cheater's Bowl, an interface to
      collect user traces as they try to answer difficult questions.
      Players see a question\protect\footnotemark (top), can search for information (left),
      view information (center), and give their answer (top) with associated
      evidence (right).}
    \label{fig:interface}
\end{figure*}

Some of the online discussions around online cheating revealed that
some people actually enjoyed doing these quick dives for information.
Thus, a goal of this paper is to see if we could (1)
sublimate these urges into something more wholesome, (2) gather useful
data to understand human expert search, and---a less scientific
question---(3) see who is the best at cheating in trivia competitions.
To answer these questions, we create a gamified interface
(Figure~\ref{fig:interface})---which we call Cheater's Bowl---to help
players find answers.

Because players come from the trivia playing
community, they know substantially more about the topics than, say, crowdworkers.
This puts them closer to the ``expert'' category discussed by
\citet{allen-91:topic}.
We use Quizbowl questions~\cite[\qb]{boyd-graber-12}, where each
question is a sequence of clues with the same answer of decreasing
difficulty (as decided by a human editor).
We also include questions from \hotpot{}~\cite{yang-etal-2018-hotpotqa}, a
popular dataset for multi-hop question answering.
We filter the questions in two ways to ensure that both humans and
computers are challenged.
First, we discard all but the two hardest clues, which should be difficult
for most humans (even our experienced player base).
Second, we try to answer all of these questions with current
state-of-the-art BERT-based model on the
data~\cite{rodriguezQuizbowlCaseIncremental2021} with a single hop.
If the model can answer the question with any number of clues, we
exclude it from the questions set used in the data collection.

%




\footnotetext{The figure is a screenshot to illustrate the interface during player training. The question is not a part of the experiment data.}



\subsection{Game Interface}

The player is presented with a question, initially with only one clue.
To start searching, the players have the option of typing their own queries in the search box, or clicking on a model-suggested query (from \abr{irrr} or \abr{golden}).
The search engine returns results from two different retrievers:
\bmt{},
a sparse index based on lexical similarity; and
Dense Passage Retrieval~\cite[\abr{dpr}]{karpukhin-etal-2020-dense},
which uses dense vector embeddings of passages.
Both retrievers index and return paragraphs from Wikipedia pages.
We use ElasticSearch~\cite{gormley2015elasticsearch} to implement
\bmt{}, and for \abr{dpr}, we directly use a pretrained model.

Both retrievers return the top passages by cosine similarity. Players can click on the
Wikipedia page titles of the passages; the full Wikipedia page is
then shown in the main document display with the passage highlighted.

The popup tooltip provides shortcuts to directly query the search
engine from highlighted text, record it as evidence, or submit it as
an answer.
Players are encouraged to highlight and record text as
evidence if it is helpful for them to find the answer.
Even if a player does not record any evidence, the paragraphs a player
reads are automatically recorded as evidence.

If the player finds the question difficult to answer, they are
free to skip the question or ask the system to reveal another
clue.\footnote{This only applies to \qb{} questions.}

\paragraph{Human-computer collaboration.}

In addition to the queries from \abr{golden} and \abr{irrr}, players also
see \abr{irrr}'s answers.
Players can directly answer the question with suggested
answers (but are encouraged to find evidence to back it up).

\paragraph{Scoring system.}
Our goal is to create an interface that is both fun and useful for
collecting relevant information.
%
Players are rewarded for having the highest score, and they earn points by:
(1) answering more questions, as each question adds to their score;
(2) answering questions correctly (100 points for each correct
answer);
(3) answering quickly, as the possible points decrease with a timer
(four minutes for \qb{} questions, three for \hotpot{});
(4) answering with fewer clues, as it makes the question easier (each
clue removes ten points);
(5) recording more evidence. Each piece of recorded evidence is awarded ten
points.


\subsection{The Player Community}
We recruit thirty-one players from the trivia community who played the game
over the course of the week.
The top player answered 895 questions, and thirteen players answered at least forty questions.
After filtering out empty answers and repeated submissions of the same
player on the same question, we collect 2545
questions-answer pairs from \qb{} of which 1428 were correctly
answered (56\%), as well as 315 pairs from
\hotpot{}, of which 225 were correctly answered (71.43\%).

\subsection{A Question Answering Example}

To see how a player might answer the question with our interface, 
we present a question-answering example with corresponding player actions (Figure~\ref{fig:qa_action}).
Answering this question requires figuring out who the main speaker was
(Prem Rawat) and then figuring out his nationality to get to the final
answer, \underline{India}.
The player answers the question with two hops: first to ``Millennium
'73'' and then to ``Prem Rawat'', finally uses commonsense
reasoning to answer ``India''.
Player actions and seen paragraphs are automatically recorded through
the process.

\begin{figure*}[t]
    \begin{center}
      \noindent\fbox{
        \parbox{.98\linewidth}{
          \small
        {\bf Question:} ``A 15-year-old
religious leader originally from this country spoke at a highly
anticipated event at which it was predicted that the Astrodome would
levitate; that event was Millennium '73''. {\bf Answer}: ``India''.\\
    (1) Query $q_0=$``Millennium '73'' \textcolor{deepblue}{(Substring of question)}\\
    (2) Select and read the Wikipedia page: ``Millennium '73''. Manually record evidence $d_1=$`` It featured Prem Rawat, then known as Guru Maharaj Ji, a 15-year-old guru and the leader of a fast-growing new religious movement.'' \\
    (3) Query $q_1=$``Prem Rawat'' \textcolor{deepblue}{(Substring from evidence $d_1$)} \\
    (4) Select and read the Wikipedia page: ``Prem Rawat''. Manually record evidence $d_2=$``Prem Pal Singh Rawat is the youngest son of Hans Ram Singh Rawat, an Indian guru.'' \\
    (5) Answer $a=$``India'' \textcolor{deepblue}{(Derived from evidence $d_2$)}
}}
\end{center}

\caption{An example of player actions for question answering with
  action path $A=(q_0, \mathcal{E}_1, q_1, \mathcal{E}_2, a)$, where
  $\mathcal{E}_1=\{d_1\}$ and $\mathcal{E}_2=\{d_2\}$.  The player
  uses substrings from the question and evidence as queries, and
  derives the final answer from evidence.  We highlight the source of
  actions in blue.  Our goal is to use these interactions to better
  understand computers' question answering.}

    \label{fig:qa_action}
\end{figure*}

%% file: sections/40-analysis.tex
\section{Human vs. Computer Search Strategies}\label{analysis}

This section compares and contrasts how computers and humans search
for information.

\subsection{Strategies in Common}

Agents search Wikipedia using
text queries, process results, and give an answer.
Both humans and computers often
create queries from the question: 83.05\% of human queries have at
least one word from the question, while 84.61\% of \abr{golden} queries and
99.75\% of \abr{irrr} queries do.
Both use terms from the evidence they find to create new queries:
14.47\% of human queries have at least one word from retrieved evidence, while 19.13\% of \abr{golden} and 28.30\% of \abr{irrr} queries
do.
Both reformulate their queries based on evidence to uncover new information~\cite{xiongAnsweringComplexOpenDomain2020}.

%
  

\subsection{Where Strategies Diverge}\label{sec:strategy_diff}


\paragraph{Humans use fewer but more effective keywords.}

\begin{figure}[t]
    \centering
    \includegraphics[width=\linewidth]{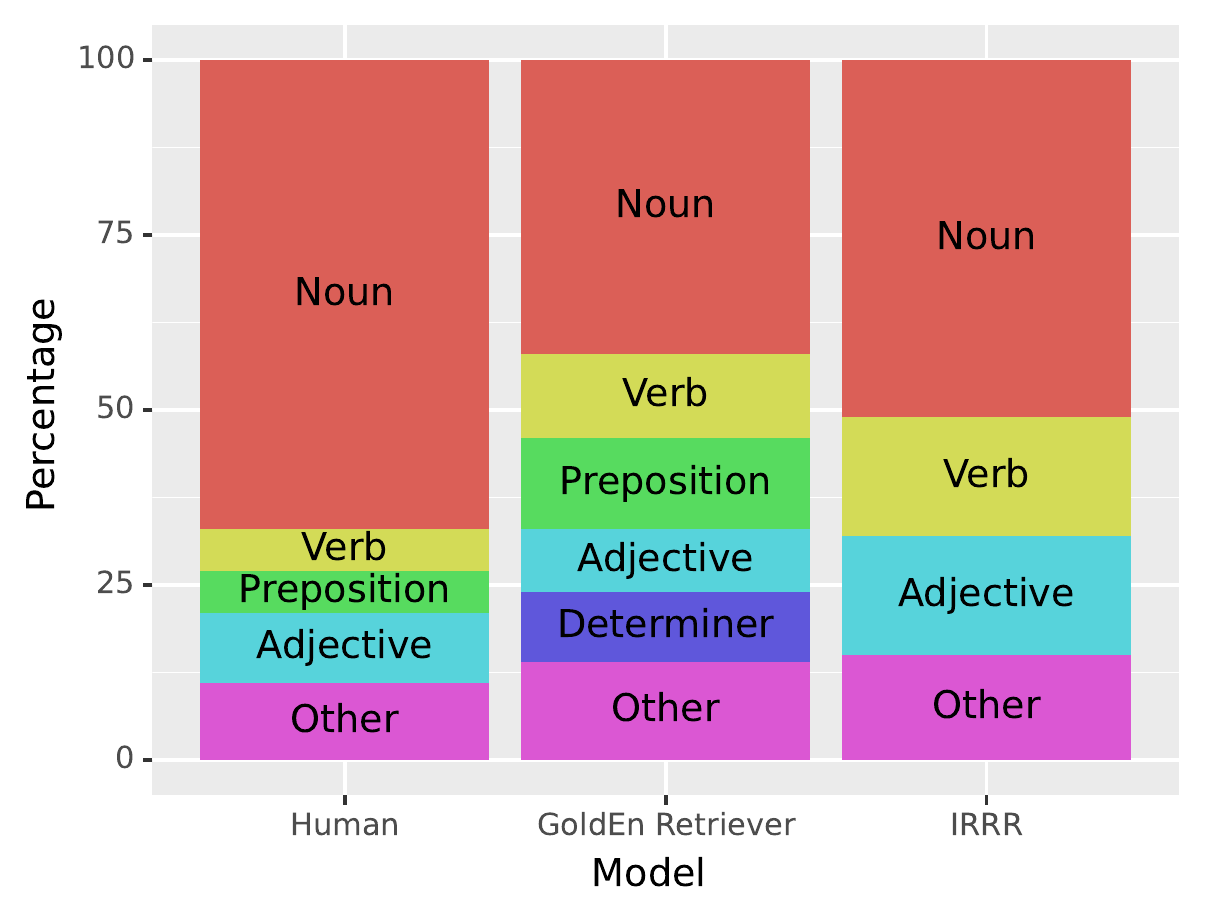}
    \caption{Proportion of different part-of-speech tags used in
      queries detected by the the Natural Language
      Toolkit~\cite[\abr{nltk}]{nltk}.  Humans focus more on nouns,
      eschewing determiners for the sake of brevity.  On this basis,
      \abr{irrr} is the more ``human'' of the two computer agents.}
    \label{fig:postag}
\end{figure}

The most salient difference between human and computer queries is that
human queries are shorter.
Human queries contain 2.67 words on average (standard deviation of 2.46);
while \abr{golden} retriever's contain 7.03$\pm6.84$ words, and \abr{irrr}'s have 12.76$\pm5.64$.
Human queries focus on proper nouns and short phrases (Figure~\ref{fig:postag}).
Humans tend to select the most
specialized term---e.g., the entity most likely to have a
comprehensive Wikipedia page---which requires world knowledge (Table~\ref{table:keywords}).
In contrast to humans' desire for precision, models prefer
recall with as many keywords as possible, hoping that it retrieves
something useful for the next hop.

\begin{table*}[]
\centering
\small
\begin{tabular}{p{6.4cm}p{1.9cm}p{4.3cm}p{2.3cm}}
\toprule
\multicolumn{1}{c}{\multirow{2}{*}{Question and answer}} &
  \multicolumn{3}{c}{First query} \\ \cmidrule(l){2-4} 
\multicolumn{1}{c}{} &
  \multicolumn{1}{c}{Player} &
  \multicolumn{1}{c}{\abr{irrr}} &
  \multicolumn{1}{c}{\abr{golden} retriever} \\ \midrule
Q: Evans et al. developed bisoxazoline complexes of this element to catalyze enantioselective Diels-Alder reactions. A: Copper &
  Evans \textbf{auxiliary} &
  Evans \textbf{et al. developed bisoxazoline complexes element catalyze enantioselective Diels-Alder reactions} &
  Evans \textbf{et al.} \\ \midrule
Q: This quantity's name is used to describe situations in which there exists a frame of reference such that two given events could have happened at the same location. A: time &
  frame of reference \textbf{same} location &
  \textbf{quantity's name used describe situations exists} frame reference \textbf{two given events could happened} location &
  \textbf{quantity's name is used to describe situations} \\ \midrule
Q: Discovered in 1886 by Clemens Winkler, this element is used in glass in infrared optical devices, its oxide has been used in medicine, and its dioxide is used to produce glass with a high index of refraction. A: Germanium &
  Clemens Winkler &
  \textbf{Discovered 1886 Clemens} Winkler element \textbf{used glass infrared optical devices oxide used dioxide used glass high index refraction} &
  \textbf{Discovered in 1886 by} Clemens Winkler \\ \midrule
Q: In ruling on these documents, the Court held that the ''heavy presumption'' against prior restraint was not overcome. A: Pentagon Papers &
  heavy presumption prior restraint &
  \textbf{ruling documents Court held} ''heavy presumption'' \textbf{against prior} restraint overcome &
  \textbf{ruling on these documents, the Court} \\ \midrule
Q: One of this director's films introduced the cheery song ``High Hopes,'' while another describes the presidential campaign of Grant Matthews. A: Frank Capra &
  high hopes song &
  \textbf{One director's films introduced cheery} song ``High Hopes '' \textbf{describes presidential campaign Grant Matthews} &
  \textbf{director's films introduced the cheery} song ``High Hopes,'' \\ \bottomrule
\end{tabular}
\caption{The first query for each question from different
  agents. Computers' words that are distinct from humans' are in {\bf bold}.
   Human queries contain fewer keywords and focus more on
  precision, while computer queries focus more on recall.}
\label{table:keywords}
\end{table*}

\paragraph{Humans use world knowledge to narrow search results.}
Unlike computers, humans sometimes use words that are not in the
question or in evidence: 16.30\% of queries have terms in neither
evidence nor question text (compared to 0\% for both
computer methods).
In the first example in Table~\ref{table:keywords}, the player's first query
is derived from the question but adds ``auxiliary'', recognizing that
``treating'' a compound makes it an auxiliary in the reaction. 
Players also reported in the feedback survey that adding a subject
category (for example, adding ``chemist'' when querying a person in
chemical-related questions) can be useful for restricting the search
results.
Although there are cases when players directly query terms closely
related to the answer, in most cases, people use common sense to narrow
the search scope or use domain-specific knowledge they have learned
from previous searches.
These patterns could be potentially learned by \qa{} models, as we
discuss further in Section~\ref{model}.


\paragraph{Dynamic query refinement and abandonment.}

Although both humans and computers reformulate queries
as a search strategy, how humans reform their queries is more
advanced.
Not all retrieved documents help lead to the answer: some are
irrelevant, and some are even misleading.
In cases when human agents have not found any helpful information from
the documents~$\mathcal{D}_i$ retrieved by query~$q_i$ or when they
are confused and unsure, the human agent does not need to use a
document from $\mathcal{D}_{i+1}$ for making new queries.
If that happens, they ignore the useless evidence, 
i.e. $\mathcal{E}_{i+1} = \emptyset$.
Instead, they write a new query $q_{i+1}$ by adding more constraint
words and deleting distracting terms from $q_i$ to restrict the
search scope or abandon $q_i$ and write a completely new query.
\citet{russell_2019_parrotfish} describes querying ``stoplight
parrotfish sand'' to uncover the relationship between parrotfish and
geology, however, the results are too mixed to be useful. He then
modifies the query to ``parrotfish sand'', which yields good results.

However, for \abr{golden} retriever and \abr{irrr}, even when irrelevant documents
are retrieved from a bad query~$q_{i}$, the model is compelled to
select some $d_{i+1}\in \mathcal{D}_{i+1}$ as evidence, append to the
reasoning path, and generate subsequent queries accordingly.
As an example, to answer the question \begin{quote}
He lost the presidential
election in 1930, which was not good enough for him as later that year
he seized power at the head of an army-backed coup. (Answer:
Getúlio Vargas (a Brazilian president))
\end{quote}
\abr{irrr} queries ``lost presidential election 1930 year seized power
head army backed coup'' but an article about Brazil is not in the
returned results.  \abr{irrr} then appends a paragraph from the
irrelevant page about the Nigerian president ``Olusegun Obasanjo'' to the reasoning path, leading to the next query ``lost
presidential election 1930 later year seized power head army backed
coup Olusegun Obasanjo'' which prevents finding a relevant Brazilian
page.

\paragraph{Multiple search chains.}

A search chain is searches~$(q_s, q_{s+1}, q_{s+2}, \cdots, q_t)$.
New searches depend on old ones, either because a subsequent search~$q_{i+1}$
refines a previous search---$q_i$ or~$q_{i+1}$, for example---integrates
evidence~$\mathcal{E}_{i+1}$ retrieved from~$q_i$.
A search chain breaks when $q_i$ is abandoned and $q_{i+1}$ is
unrelated to previous evidence or queries.
While existing computer agents can only use a
single search chain, human agents use multiple search chains,
either pre-planned parallel search chains that focus on different
perspectives of the question, or starting a new one if previous chains
fail to lead to the answer.  When answering the question
\begin{quote}
This modern-day country was once ruled by renegade Janissaries known
as dahije, who massacred this country's elite, known as knez, in
1804. (Answer: ``Serbia'') 
\end{quote}
the player first makes a query about the mentioned title ``knez'', and
next queries ``Knyaz'', which is a substring of the evidence retrieved
by the first query.
However, these queries fail to retrieve useful results since
``knez'' and ``Knyaz'' are common Slavic titles.
The player then abandons this search chain and starts a new one with
the query ``dahije'', which allows the player to retrieve the
Wikipedia page ``Dahije'' that includes the answer ``Serbia''.


\paragraph{Swapping Engines.}
The \textit{Joy of Search} is replete with searches over different
sources: Google, Google Scholar,
Google Earth, etc.
While we only give players access to Wikipedia, we allow players to
switch between ElasticSearch and \abr{dpr}.
In contrast to multi-hop systems which typically use trained, dense
retrievers, players prefer ElasticSearch (87\% of queries) over
\abr{dpr}.
Some of this is probably familiarity: most public-facing search
engines (including Wikipedia) are term-based retrievers.
In the post-task survey, players prefer ElasticSearch because it is
most useful when looking for an exact Wikipedia page---the specific
Wikipedia page always ranked top among search results.
It is also helpful for checking answers: they often query an answer
candidate, which helps boost their answer accuracy.
ElasticSearch---given its predictability---is better for this specific strategy.

\paragraph{Beyond a Bag of Words.}
However, this is not always the case; when humans do use \abr{dpr},
they adapt their query styles for better retrieval.
Some players report that they could find their desired results with
natural language queries when using \abr{dpr}.
Those queries usually come from longer sequences in the question or
evidence. For example, when answering the question
\begin{quote}
Mathilda Loisel goes into debt to replace paste replicas of these
gemstones, one of which is ``As Big as the Ritz'' in an F. Scott
Fitzgerald short story.  (Answer: ``Diamond'')
\end{quote}
the player queries ``As Big as the Ritz'' in an F. Scott Fitzgerald
short story.'' with \abr{dpr}, which retrieves the Wikipedia page
``The Diamond as Big as the Ritz'' with the answer.

Players also report searching Google with natural language queries
when finding answers to open-ended questions with various options,
e.g., ``How often should I wash my car?''.
In these scenarios, humans
may search for relatively vague queries and synthesize an answer from
multiple retrieval results.
{\small Web}\abr{gpt}~\cite{nakanoWebGPTBrowserassistedQuestionanswering2021}
explores a similar setting by training
\abr{gpt}-3~\cite{brownLanguageModelsAre2020a} to search queries in natural
language, aggregate information from multiple web pages and answer
open-ended questions.  Due to the limitation of Cheater's Bowl where most \qb{} answers are Wikipedia titles~\cite{rodriguezQuizbowlCaseIncremental2021},
agents do not need this more flexible setup.







%% file: sections/50-model.tex
\section{Existing Models and Future Design}\label{model}

\begin{figure}[t]
    \centering
    \includegraphics[width=\linewidth]{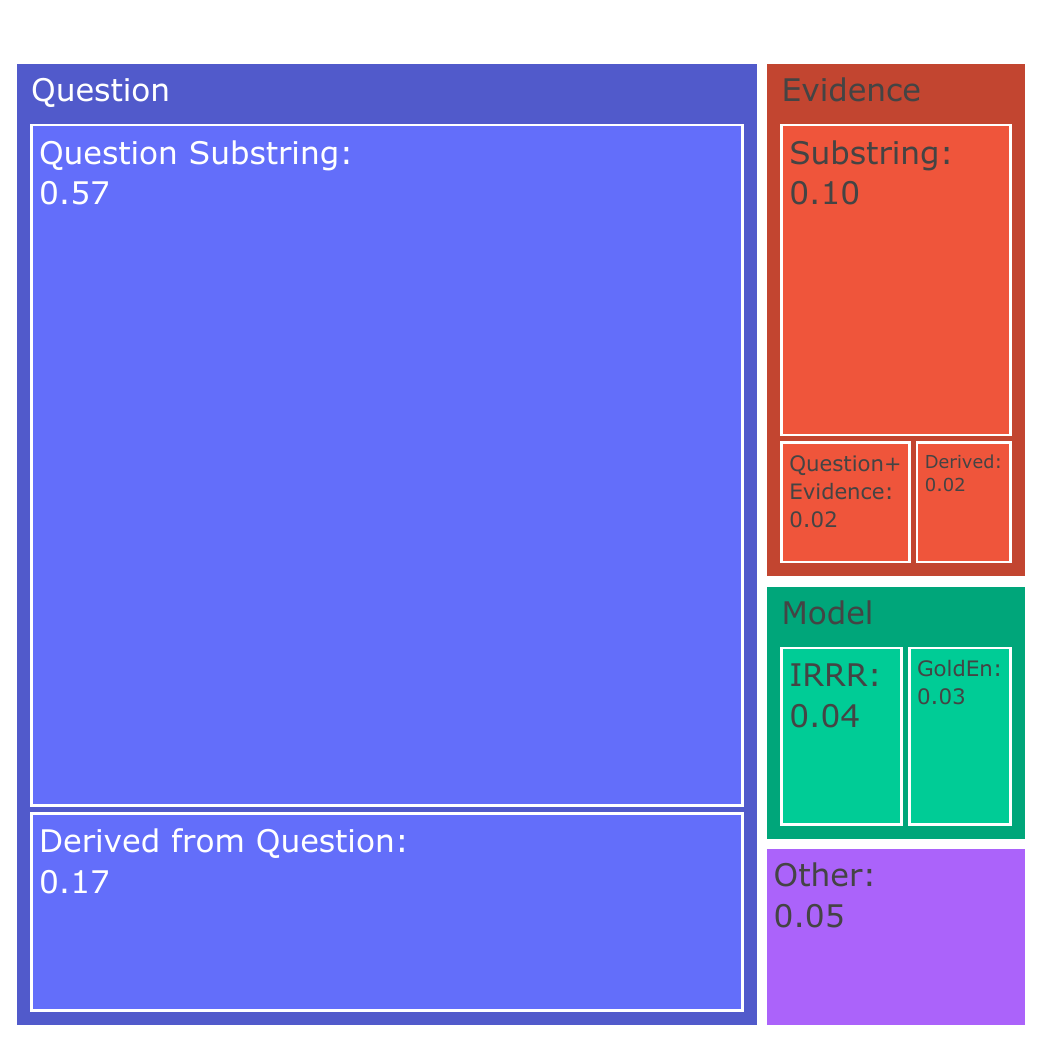}
    \caption{Users need to find the answer to a question but have
      several sources that might inspire their queries: the original
      question, evidence, or models.  This Treemap shows the source
      for their questions (area corresponds to frequency).  Only a
      small proportion of queries are from \qa{} models' suggestions.}
    \label{fig:query_origin}
\end{figure}

\begin{figure}[ht]
    \centering
    \includegraphics[width=\linewidth]{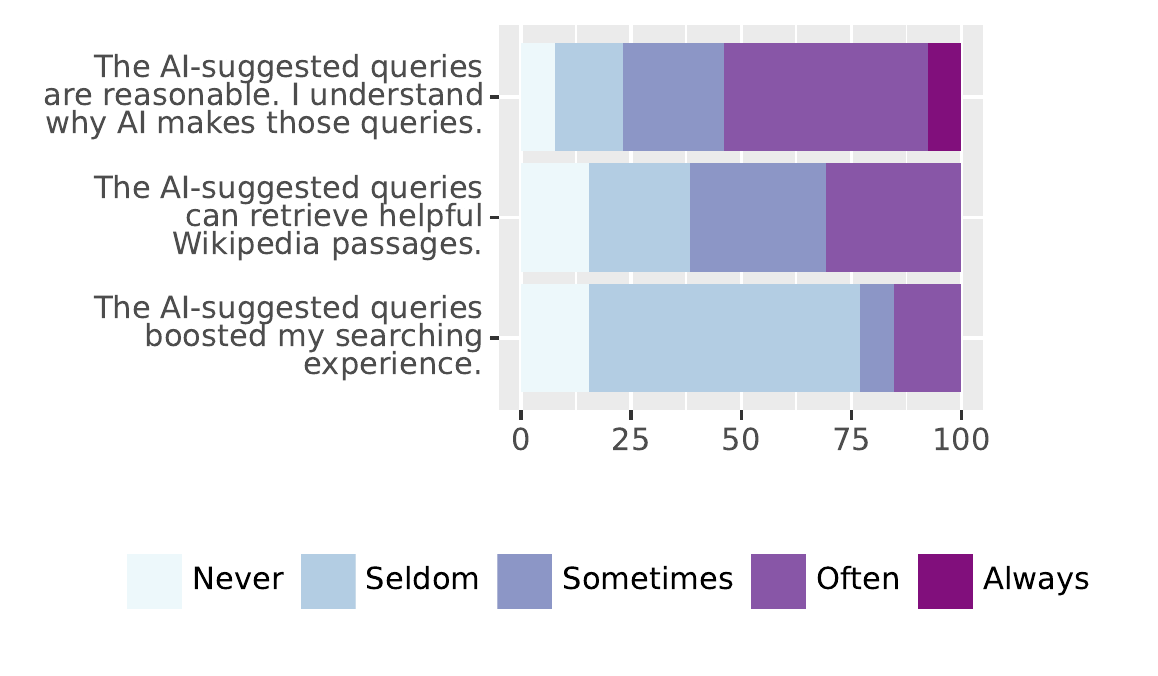}
    \caption{From our post-survey, players' feedback for queries suggested by \qa{} models.
      Although most players understand why models make those queries,
      players doubt the utility in improving the search experience.}
    \label{fig:AI_feedback}
\end{figure}

Although we present queries suggested by state-of-the-art multi-hop
\qa{} models to players, players would rather write their own queries
(Figure~\ref{fig:query_origin}).
Most players understand why \qa{} models query the way they do
(Figure~\ref{fig:AI_feedback}) and agree that queries retrieve helpful
results, but players question the utility.
%
This is an intrinsic difference between humans and models: human queries strive for a
``direct hit'' with two to three search results, as
\cite{JANSEN2000207} find that humans only access
results on the first page, typically the first few results. In
contrast, verbose model queries hope search results contain
\emph{something} helpful---it does not mind reading through a dozen
search results.
Another reason might be that \qa{} models are worse than humans: for
\qb{} questions randomly given to players, 56.58\% of the questions
are correctly answered by players, while only 44.21\% are correctly
answered by \abr{irrr}.\footnote{For questions randomly sampled from
  \hotpot{}, human accuracy (71.43\%) is slightly lower than
  \abr{irrr}'s 79.02\%. We consider this to be due to the synthetic
  construction of the \hotpot{} dataset: it lends itself to straightforward
  searches.  \qb{} better discriminates (by design) between agents'
  ability.}
%

\subsection{Improve Existing Models with Human Actions}

Though \qa{} models fail to help humans advance their searches,
could the accuracy of the \qa{} models increase if we replace 
computer queries with humans'?
We compare how well \abr{irrr} performs under two settings: querying and
answering from scratch ({\bf scratch}) v.s. initializing the model's reasoning
path from the human reasoning path ({\bf init from human}).


To convert human queries into \abr{irrr}'s format, given the full action path $A=\left(q_{0},
\mathcal{E}_{1}, q_{1}, \mathcal{E}_{2}, \cdots,q_{k-1},
\mathcal{E}_k, a\right)$ of question $Q$, for each $0\leq j\leq k-1$,
we trim the action path that ends with query~$q_j$ to form a partial
human action path~$A_j=\left(q_{0}, \mathcal{E}_{1}, q_{1},
\mathcal{E}_{2}, \cdots, q_j\right)$.
We initialize the reasoning path $R$ with $R=(Q)$.
Then, we extend the reasoning path with documents from the user.
Because \abr{IRRR} can only look at one document at a time, we need to
decide which document to append to the reasoning path~$R$.
For each $\mathcal{E}_i$ ($1\leq i\leq j$) in action path $A_j$,
if~$\mathcal{E}_i\neq \emptyset$, we append the most crucial document
$d_i\in \mathcal{E}_i$ in this order: 1) source of
player answer 2) source of some query 3) manually recorded by the
player as evidence, since they are more likely to lead the model to
generate human-like queries and answers.
%
We consider the converted human
reasoning path $R_l=(Q, d_1, d_2, \cdots, d_l)$ to be the reasoning
path of reasoning step $l$, where $l\leq j$ since there might be empty
$\mathcal{E}_i$. Note that as the special case, we consider $R_0 = (Q)$ and $A_0=(q_0)$.


%
We compare the two settings on the question set $\mathcal{Q}_l$, which
is the set of questions where partial human actions $A_j$ could be
converted to a human reasoning path at reasoning step $l$
($0\leq l\leq 2$).
Obviously $\mathcal{Q}_2 \subseteq
\mathcal{Q}_1 \subseteq \mathcal{Q}_0$. We have converted $
|\mathcal{Q}_0| = 1122, |\mathcal{Q}_1| =462, |\mathcal{Q}_2| =195 $
questions in total.  The difficulty of questions in $\mathcal{Q}_2$
is, in general, greater than questions $\mathcal{Q}_0$ since humans
use at least three queries for answering the questions in
$\mathcal{Q}_2$, while using at least one query for $\mathcal{Q}_0$.

Initializing from human actions significantly improves the accuracy of
the final answer (Table~\ref{table:irrr_acc}), outperforming querying
from scratch by 10.26\% for questions in $\mathcal{Q}_2$.
The human queries can unlock reasoning paths that make previously
unanswerable questions answerable within three steps.
While humans cannot get much from computer queries, the reverse is
certainly true.
We further qualitatively analyze why human actions are helpful
to models.

\begin{table}[]
\centering
\begin{tabular}{@{}ccc@{}}
\toprule
\textbf{Questions} & \textbf{Scratch} & \textbf{Init from human} \\ \midrule
$\mathcal{Q}_0$ & 44.21\% & 50.45\% \\
$\mathcal{Q}_1$ & 38.10\% & 42.42\% \\
$\mathcal{Q}_2$ & 27.69\% & 37.95\% \\ \bottomrule
\end{tabular}
\caption{Compared to querying from scratch, \abr{irrr} answer accuracy
  greatly increases after initializing from human actions, suggesting
  models benefit from humans' insights.}
\label{table:irrr_acc}
\end{table}



\paragraph{Better selection of keywords.}

For questions where \abr{irrr} answers correctly with human
initialization but fails alone, 91.48\% of the first queries are
substrings or derived from the question.
Models select more keywords (Section~\ref{sec:strategy_diff});
however, this strategy might fail when the retrieval results are too
diffuse.
In the last example from Table~\ref{table:keywords}, the first
\abr{irrr} query retrieves weakly related documents, and \abr{irrr}
appends a paragraph from ``Cultural impact of the Beatles'' to the
reasoning path.
Since \abr{irrr} can only use a single search chain, the second and
the third query follow previous evidence and retrieves more irrelevant
documents.
In comparison, the player query ``high hopes song'' allows
\abr{irrr} to find ``High Hopes (Frank Sinatra song)'' and use it as
evidence. That paragraph contains key information---the film \textit{A
  Hole in the Head}---which unlocks the film's director, Frank Capra.

\paragraph{World Knowledge.}
A small proportion of human queries ``improves'' the model accuracy
because it directly includes the answer or shortcuts to the answer. As
an example, the first human query for the question
\begin{quote}
  The first one of these to be directly observed was obtained by the
  solution of TBF in an antimony-based superacid.
\end{quote}
is ``George Olah'', a Hungarian-American chemist associated with
``superacids''.
\abr{irrr} uses this shortcut to find the answer ``carbocations'' on
the Wikipedia page ``George Andrew Olah''.

\subsection{Design Suggestions for Future Models}

Based on the strategic differences between human and \qa{} models, we
propose improvements for future query-driven open-domain \abr{qa}
models.

\paragraph{Retriever-Aware Queries.}
The model should be able to interact with the retrieval system,
dynamically refine imperfect queries based on retrieval results, and
abandon search chains that cannot lead to the answer.
Queries can be refined by deleting and adding words, using search
operators~\cite{adolphsBoostingSearchEngines2021}, or adding masks to
tokens for dense queries~\cite{zhang-etal-2021-extract-integrate}.
Query refinement can be trained through reinforcement learning~\cite{adolphsBoostingSearchEngines2021} or supervised learning from a synthetic query reformulation dataset~\cite{Adolphs_Decoding}.
If retrieval results are irrelevant to the question, the model should
discard the results: $\mathcal{E}_i=\emptyset$, avoiding the
introduction of noise for future query generation.
Models should be able to dynamically select search engines and specify
search sources suitable for each query.

\paragraph{Incorporate Common Sense and World Knowledge.}
Instead of using substrings from questions and previous
evidence as queries, the model should add 
words and terms to queries just like humans
do, either via templates, or using a
generative language model~\cite{Li2022}.
Other methods for incorporating common sense
and world knowledge include accessing an external knowledge
base~\cite{LUNAR, harabagiu_finding_nodate} and reasoning over
knowledge graphs~\cite{lin-etal-2019-kagnet,zhang2022greaselm}.

\paragraph{Check Your Work.}
Models should explicitly check the correctness of candidate answers.
A simple yet effective strategy humans use is to directly query the
candidate answer and see whether it can retrieve documents related to
the question.
Previous research also explores answer validation through abduction~\cite{harabagiu_finding_nodate} and via Web
information~\cite{magnini-etal-2002-right}.

A model that satisfies the above design principles could be
implemented using reinforcement learning with a well-defined
environment and reward function.
Such a system would provide flexibility---enabling dynamic query
refinement~\cite{Huebscher2022ZeroShotRW} and abandonment---which are
not supported by traditional \qa{} systems.
This direction would also solve more complex \qa{} tasks that require
planning and balancing, e.g., answering incremental
questions~\cite{rodriguezQuizbowlCaseIncremental2021} with fewest clues and fewest searches.
The contributions in this paper make this possible; for example, a reinforcement learning agent
could be trained from our data by behavior cloning.

%% file: sections/60-related.tex
\section{Related Work}


\paragraph{Human Use of Search Engines.}
Our work is similar to previous research that analyzes the behavior of humans using search engines.
\cite{oday_vicki} discover that it is crucial to reuse the results from the previous searches to address the information need.
\cite{lau_horvitz} evaluate the logs of the Excite search engine and find that each information goal requires 3.27 queries on average.
\cite{jansen_2009, huang_jeff} find that contextual query refinement is a widely used strategy. Queries are refined by incorporating background information and evidence from past search results, which usually include examining result titles and snippets.
Our work provides many of the same features as these previous papers but adds neural models to retrieve passages, suggest queries, and extract answers. Our analysis focuses on comparing human and computer search strategies and how they may benefit each other. 
In addition, our task gamifies the search task using the unique structure of \qb{} questions, which is intended to make the task more challenging and fun.


\paragraph{Question Answering Agents.}
Previous work has explored agents that issue interpretable text-based queries to a search engine to answer questions.  
\abr{golden} retriever~\cite{qi-etal-2019-answering} generates a query by selecting a span from the reasoning path, and \abr{irrr}~\cite{qi-etal-2021-answering} further advances the \abr{golden} retriever by allowing queries to be any subsequence of the reasoning path.
\citet{adolphsBoostingSearchEngines2021} train an agent using reinforcement learning
to interact with a retriever using a set of search operators.
%
WebGPT~\cite{nakanoWebGPTBrowserassistedQuestionanswering2021} is a large language model based on GPT-3~\cite{brownLanguageModelsAre2020a} that searches queries in natural language and aggregates information from multiple web pages to answer \emph{open-ended} questions.

\jbgcomment{Please expand this related work and remove citation dumps; I suspect there are missing references from 2022 that should probably add}

\paragraph{Alternative Models.}\label{sec:alt}
While our work only compares human search strategies with computer systems that answer questions by searching text-based queries, modern retrievers directly compose vector queries~\cite{karpukhin-etal-2020-dense, xiongAnsweringComplexOpenDomain2020, zhao-etal-2021-multi-step}, hop through different documents by following structured links~\cite{asaiLearningRetrieveReasoning2019, zhaoTransformerXHMultievidenceReasoning2020}, or resolve coreference~\cite{chenMultihopQuestionAnswering2019}.
However, we consider that vector-based queries are confusing black boxes for human players. Thus, computer systems using vector-based queries could hardly collaborate with humans. 
Players say they use Wikipedia links to directly jump to other Wikipedia pages. Thus, following these structured links or resolving coreference could be modeled by query-generation systems: if a user clicks on a Wikipedia link, that could be a part of the next query.


%% file: sections/70-conclusion.tex
\section{Conclusion}

Open-domain and multi-hop \abr{qa} is an important problem for both
humans and computers.  To compare how humans and computers search and
answer complex questions, our interface collects human question answering data as agents search with traditional and neural search
engines alongside question answering models that suggest queries and
answers.
Humans often use shorter queries, apply dynamic search chains, and use
world knowledge.
Future \abr{qa} models should have the ability to generate novel
queries, ``discard'' irrelevant results, and explicitly check 
answers.
Moreover, computer agents for \abr{qa} should also be able to use
diverse retrievers to find evidence to answer
questions, learning from the insights found in human data.
With an agent trained on our data, we could have the ``best of both
worlds'' to combine the ingenuity and tacit knowledge of humans with
an indefatigable agent with access to all the world's information.









%% file: sections/acks.tex
\section*{Acknowledgements}
We thank Michelle Yuan, Shi Feng, Chenglei Si and the anonymous
reviewers for their insightful feedback.
We thank Tsinghua University (Wanrong He) and the National Institute
of Standards and Technology (Andrew Mao) for fellowships that
supported the authors' research experience.
Any opinions, findings, and conclusions or recommendations expressed
in this material are those of the researchers and do not necessarily
reflect the views of the funders.
We thank the Cheater's Bowl participants for supporting this work by
providing their search session data.  Boyd-Graber is supported by NSF
Grant IIS-1822494 and by \abr{odni}, \abr{iarpa}, via the \abr{better}
Program contract \#2019-19051600005.
The views and conclusions contained herein are those of the authors
and should not be interpreted as necessarily representing the official
policies, either expressed or implied, of \abr{odni}, \abr{iarpa}, or
the \abr{us} Government. The \abr{us} Government is authorized to
reproduce and distribute reprints for governmental purposes
notwithstanding any copyright annotation therein.

\section*{Limitations}
The first limitation of this work is that we only provide Wikipedia as the single source for information retrieval because Wikipedia is the common retrieval source used in open-domain \qa{} models; hence we failed to directly illustrate the human behavior of searching over multiple sources. The second limitation is that for human-AI collaboration, we mainly use \abr{irrr} and \abr{golden} retriever as the representative of AI models since they are state-of-the-art multi-hop \qa{} models that generate text-based queries. \qa{} models that use different strategies could be further explored and compared with human strategies.

\section*{Ethical Concerns}

We took steps to ensure our data collection process adhered to ethical guidelines. Our study was IRB-approved. We paid players who actively participated in the gamified data collection process (\$130 for top players and \$25 for the raffle). We got feedback from the online trivia community before and after launching our game (Appendix~\ref{apx:survey}). We release our data to the public domain.

%% file: sections/apx.tex
\section{Player Feedback Survey}\label{apx:survey}
We gathered valuable feedback from our players about the data collection experiment, both to understand our human strategies, and improve our system to be more enjoyable. We sent them a questionnaire with the following questions: 

\begin{itemize}
    \item Which search engine do you prefer?
    \item How do you like these search engines?
    \item How often do you search for things from these sources? (1 to 5):
        \begin{itemize}
        \item Original question
        \item Wikipedia page (resulted from previous search)
        \item AI-suggested queries
        \item My own knowledge about the question
        \end{itemize}
    \item Please rate how much you agree with each of the statements (1 to 5):
    \begin{itemize}
        \item The AI-suggested queries boosted my searching experience. 
        \item The AI-suggested queries can retrieve helpful Wikipedia passages.
        \item The AI-suggested queries are reasonable. I understand why AI makes those queries.
    \end{itemize}
    \item Select the search strategies you have applied. (List of strategies)
    \begin{itemize}
        \item Search (multiple) keywords/specialized terms
        \item Utilize the links in Wikipedia pages, directly jump to another page
        \item Use world knowledge about the question/domain
        \item Learn domain-specific knowledge from the results, and use them in future search
        \item Add proper words to restrict the range of results (for example, the subject category like ``philosophy'', ``chemistry'', name of the topic, ...)
        \item Try name variants, e.g., Matthew C Perry $\rightarrow$ M. C. Perry
        \item Refine the previous query if it doesn’t yield any helpful results
        \item At the beginning/when unclear, make simple \& broad query (e.g. a single noun or phrase)
        \item Search candidate answer to verify its correctness
        \item Chain of searches: next query is based on previous search results
        \item Parallel searching chains: use multiple separate search chains.
        \item Search in multiple search engines.
        \item Search in multiple languages
    \end{itemize}
    \item Could you tell us more about your search strategy, and why you use it?
    \item What feature would you like to see included in this app? Is there a feature that will make finding answers easier, but we don't have it yet?
    \item Any other feedback for Cheater's Quizbowl?
\end{itemize}

Overall we received 13 responses. 

The large majority (13) of respondents preferred ElasticSearch over DPR (2), with most saying ElasticSearch better met their expectations: the Wikipedia page in their queries always ranked top. The two players who also like DPR consider DPR can retrieve what they are looking for when using natural language queries.

\begin{figure}[ht]
    \centering
    \includegraphics[width=\linewidth]{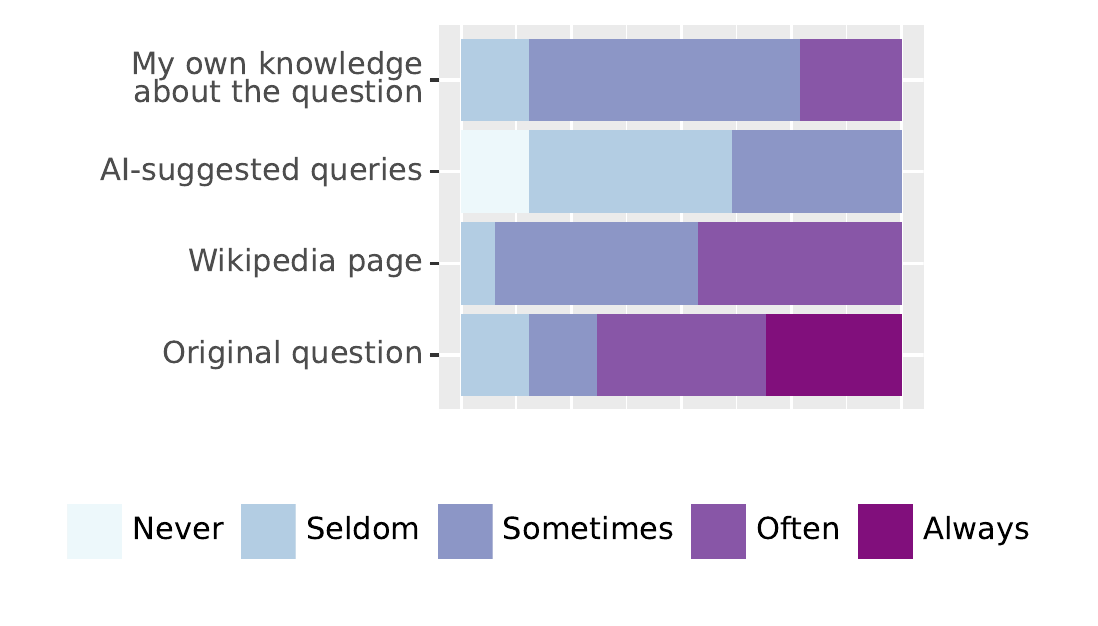}
    \caption{Source of player queries. Respondents reported that they seldomly use queries suggested by the \qa{} models.}
    \label{fig:survey_query_origin}
\end{figure}

As is shown in Figure~\ref{fig:survey_query_origin}, players mostly queries from the original question, and also from the previous retrieval results. Players seldomly use queries suggested by the \qa{} models.

Most respondents didn't find the AI suggested queries useful, but most thought they were sensible, and sometimes retrieved relevant passages (Figure~\ref{fig:AI_feedback}). 

The majority of respondents used the following strategies: clicking on Wikipedia links, refining the previous query, searching the candidate answer to validate it, creating a search chain where the next query is based on the previous passages, using multiple search chains, and using world knowledge. All strategies listed above received at least two respondents claiming that they have used it.

People also reports diverse strategies they have applied. Interesting responses includes 
\begin{quote}
    I think the inclination toward keyword search has to do with the desire for ``the'' answer rather than ``an'' answer. I definitely use natural language queries in normal searches, but usually when I am looking for a subjective answer, or a variety of options. I might google something like ``how often should I wash my car'' or ``what's the best teapot'' - questions that have possible answers, but not a single objectively correct answer. In those cases I'm happy to sort through many responses to synthesize an answer. But in Quizbowl (and especially in this case given the time/search constraints) I don't want to spend time typing a long query, or paraphrasing what's in the question, and I definitely don't want to risk getting answers that are contradictory or ambiguous. The goal is to search something specific and uniquely identifying that leads clearly to a single correct answer and keywords just seem so much safer for that goal.
\end{quote}
\begin{quote}
    Check the AI suggestions, and use one of them if they seem sensible, or type my own. Then develop it from there, based on the top results and seeing if there are any leads.
\end{quote}
\begin{quote}
    I used different strategies for different questions. I figured out quickly that the AI-generated queries were mostly not helpful for me unless they were one person's name. In those cases I found myself scanning biographical entries from the beginning and eventually getting a clue that would help me find an answer. Adding a subject category like philosophy or chemistry in the initial search was often useful. Questions about the content of literary texts and visual art were really difficult to search; I could get closer to the answer but not all the way there.
\end{quote}

\section{Implementation Detail} \label{apx:implementation}
Here we provide the implementation details for the Cheater's Bowl interface. 
\subsection{ElasticSearch}
We set up ElasticSearch with minor modifications from \cite{qi-etal-2021-answering}. We use the ElasticSearch version of 6.8.2. The index is built based on the English Wikipedia dumped on Aug 1st, 2020. We first split each Wikipedia page into paragraphs, and then index individual paragraphs (including both the text and links).

\subsection{Pretrained Models}
The pretrained \abr{irrr} model we used in our experiments can be downloaded from \url{https://nlp.stanford.edu/projects/beerqa/irrr_models.tar.gz}, and the pretrained \abr{golden} model can be downloaded through the shell script \url{https://github.com/qipeng/golden-retriever/blob/master/scripts/download_golden_retriever_models.sh}. The pretrained \abr{dpr} model we used for the search engine can be downloaded from \url{https://dl.fbaipublicfiles.com/dpr/checkpoint/retriever/single/nq/hf_bert_base.cp}.